\documentclass[11pt]{article}

\usepackage[preprint]{arxiv}

\usepackage[T1]{fontenc}
\usepackage[utf8]{inputenc}
\usepackage{lmodern}
\usepackage{microtype}
\usepackage{amsmath,amssymb,amsthm}
\usepackage{booktabs}
\usepackage{graphicx}
\usepackage{multirow}
\usepackage{url}
\usepackage{hyperref}
\usepackage{xcolor}
\usepackage{enumitem}
\usepackage{algorithm}
\usepackage{algpseudocode}

\setcitestyle{authoryear,round}
\let\cite\citep

\hypersetup{
  colorlinks=true,
  linkcolor=blue,
  citecolor=blue,
  urlcolor=blue
}

\title{Quantum Knowledge Graph: Modeling Context-Dependent Triplet Validity}

\author{
  Yao WANG$^{1,2}$ \quad
  Zixu GENG$^{3}$ \quad
  Jun YAN$^{1}$\thanks{Corresponding author.} \\
  \\
  $^{1}$HKAI-Sci, City University of Hong Kong \\
  $^{2}$Department of Automation, Tsinghua University \\
  $^{3}$Pratt School of Engineering, Duke University \\
  \texttt{ywan75@cityu.edu.hk, zg73@duke.edu, yan.jun@cityu.hk}
}

\begin{document}
\maketitle

\begin{abstract}
Knowledge graphs (KGs) are increasingly used to support large language model (LLM) reasoning, but standard triplet-based KGs treat each relation as globally valid. In many settings, whether a relation should count as evidence depends on the context. We therefore formulate triplet validity as a triplet-specific function of context and refer to this formulation as a Quantum Knowledge Graph (QKG).

We instantiate QKG in medicine using a diabetes-centered PrimeKG subgraph, whose 68,651 context-sensitive relations are further annotated with patient-group-specific constraints. We evaluate it in a reasoner--validator pipeline for medical question answering on a KG-grounded subset of MedReason containing 2,788 questions. With Haiku-4.5 as both the Reasoner and the Validator, KG-backed validation significantly improves over a no-validator baseline ($+0.61$ pp), and QKG with context matching yields the largest gain, outperforming both KG validation without context matching ($+0.79$ pp) and the no-validator baseline ($+1.40$ pp; paired McNemar, all $p<0.05$). Under a stronger validator (Qwen-3.6-Plus), the raw QKG gain over the no-validator baseline grows from $+1.40$ pp to $+5.96$ pp; the context-matching gap is non-significant ($p=0.73$) on the raw set but becomes borderline significant ($p=0.05$) after adjustment for knowledge leakage and suspicious questions, consistent with a benchmark-gold ceiling rather than a QKG limitation.

Taken together, the results support the view that the value of a KG in LLM-based clinical reasoning lies not merely in storing medically related facts, but in representing whether those facts are applicable to the specific patient context. For reproducibility and further research, we release the curated QKG datasets and source code.\footnote{\url{https://github.com/HKAI-Sci/QKG}}
\end{abstract}


\clearpage
\begin{figure}[p]
  \centering
  \includegraphics[width=\textwidth]{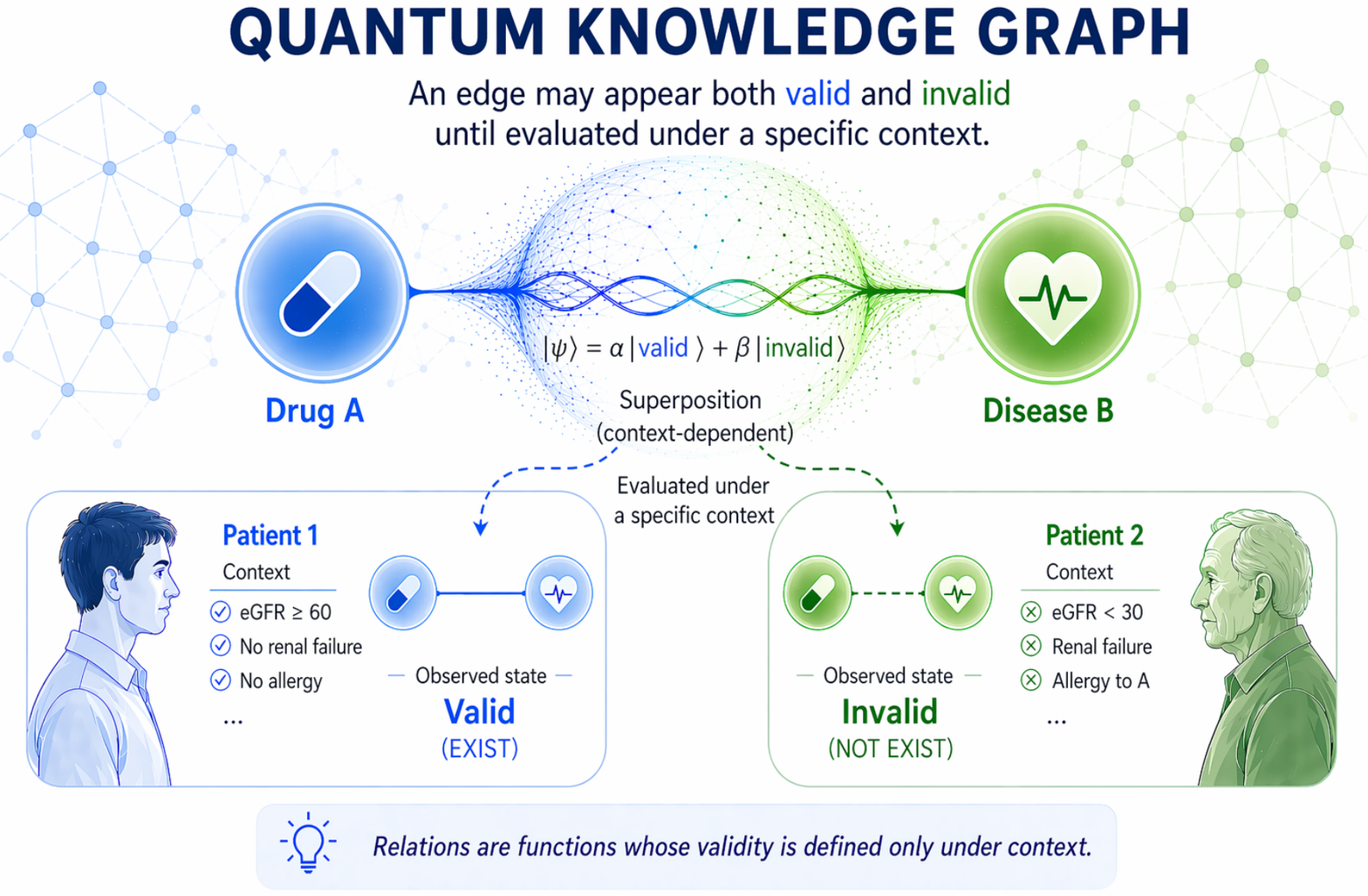}
  {\small\textbf{Graphical abstract.} Context-dependent triplet validity in a Quantum Knowledge Graph.\par}
\end{figure}
\clearpage

\section{Introduction}

% Section 1 - Paragraph 1: why research on KG is still valuable given LLM prevalence today

Large language models (LLMs) and knowledge graphs (KGs) are increasingly being developed as complementary components rather than competing paradigms. Recent work has shown that KGs can improve LLM systems by providing structured, explicit, and verifiable knowledge for retrieval, reasoning, and trustworthiness, while LLMs can in turn assist the construction, enrichment, and operational use of KGs~\cite{sui2025trustworthykgllm,wu2025medgraphrag,parovic2025llmkg}. This emerging interplay suggests that, in the LLM era, the value of KGs lies not merely in serving as retrieval resources, as in standard RAG pipelines, but in functioning as explicit and inspectable validation substrates that determine whether model-generated claims are applicable in a given context. This property is especially important in LLM-based agent systems, where generating plausible outputs is not the sole objective, and reliable performance requires verifying whether the claims underlying those outputs are supported by external evidence and context~\cite{dougrezlewis2025recv,kolli2025hybrid}.

% Section 2.1: Knowledge Graph's limitation
Conventional KGs typically represent knowledge as triples consisting of a head entity, a relation, and a tail entity. A useful way to characterize the applicability of a triplet \(\tau=(h,r,t)\) is through a context-dependent probabilistic quantity \(P(\tau \mid C)\), where \(C\) denotes the observation context. Different KG paradigms can then be viewed as different parameterizations of this quantity:
\[
P(\tau \mid C)=
\begin{cases}
\{0,1\}, & \text{conventional KG},\\
\mu_{\tau}, & \text{probabilistic KG},\\
F_{\tau}(C), & \text{triplet-specific function}.
\end{cases}
\]
Here, \(\mu_{\tau}\in[0,1]\). The first case treats triplets as universally valid or invalid, the second encodes a population-level prior, and the third allows validity to be determined by a triplet-specific function \(F_{\tau}\) that takes context \(C\) as input; in practice, \(F_{\tau}\) may be instantiated as an explicit function, a classical statistical learning model such as logistic regression or XGBoost, or an LLM.

This distinction matters because, in most real-world settings, triplet validity depends on context: \(P(\tau \mid C)\) is not constant over \(C\). Standard triplet-based KGs collapse this dependence into a binary value that records only whether a triplet holds, limiting their usefulness as validation substrates. Prior efforts have partially relaxed this assumption with structured qualifiers such as temporal scopes and hyper-relational key-value attributes~\cite{galkin2020stare,saxena2021cronkgqa}, but these still capture only selected dimensions of \(C\), whereas real-world conditions are often richer and more complex~\cite{ding2024htkg,chen2023multitq}.

% Section 2.2: KG's limitation in Medical field
This issue is especially consequential in medicine, where incorrect validation can lead to harmful conclusions. In this setting, \(P(\tau \mid C)\) is rarely a universal constant: whether a medical claim holds often depends on patient-specific context such as comorbidities, laboratory findings, disease stage, treatment history, and contraindications. Some prior work has replaced binary validity with probabilistic validity~\cite{li2020realworld}, corresponding to a population-level prior such as \(\mu_{\tau}\). While this captures aggregate uncertainty, it still does not explicitly specify the concrete contexts under which a knowledge statement should or should not be considered applicable for a particular case.

% Section 3: How to address the limitation, our theoretical approach, what is QKG
To address this limitation, we turn to triplet-specific functions and seek a practical way to implement and evaluate \(F_{\tau}(C)\). In this paper, we operationalize \(F_{\tau}(C)\) by attaching natural-language validity conditions to relations. To make this implementation scalable, triplet applicability cannot be reduced to a small set of manually engineered structured fields, because the relevant conditions are often diverse and compositional. We therefore represent these conditions as natural-language constraints, which preserve expressive flexibility while remaining compatible with LLM-based interpretation and downstream evaluation. Hereafter, we refer to the \(F_{\tau}(C)\)-based formulation of triplet validity as a Quantum Knowledge Graph (QKG)\footnote{Here, ``quantum'' refers to context-dependent validity rather than quantum-theoretic formalism.}, emphasizing that whether a knowledge statement is valid depends on the observation context in which it is evaluated.

% Section 4: our practical solution
Based on this formulation, we instantiate QKG in the medical domain by curating a graph from PrimeKG \cite{primekg} and building a validator agent that evaluates whether medical claims are supported in the patient context. We then integrate this validator into a reasoner-validator pipeline for LLM-based reasoning, and evaluate the resulting system on medical question answering using samples from MedReason \cite{wu2025medreason} whose entities are covered by the curated graph. Our experiments compare QKG-based validation against both the original KG and the no-validator baseline, and show that QKG-based validation improves system performance over both.

\begin{figure}[t]
  \centering
  \includegraphics[width=\linewidth]{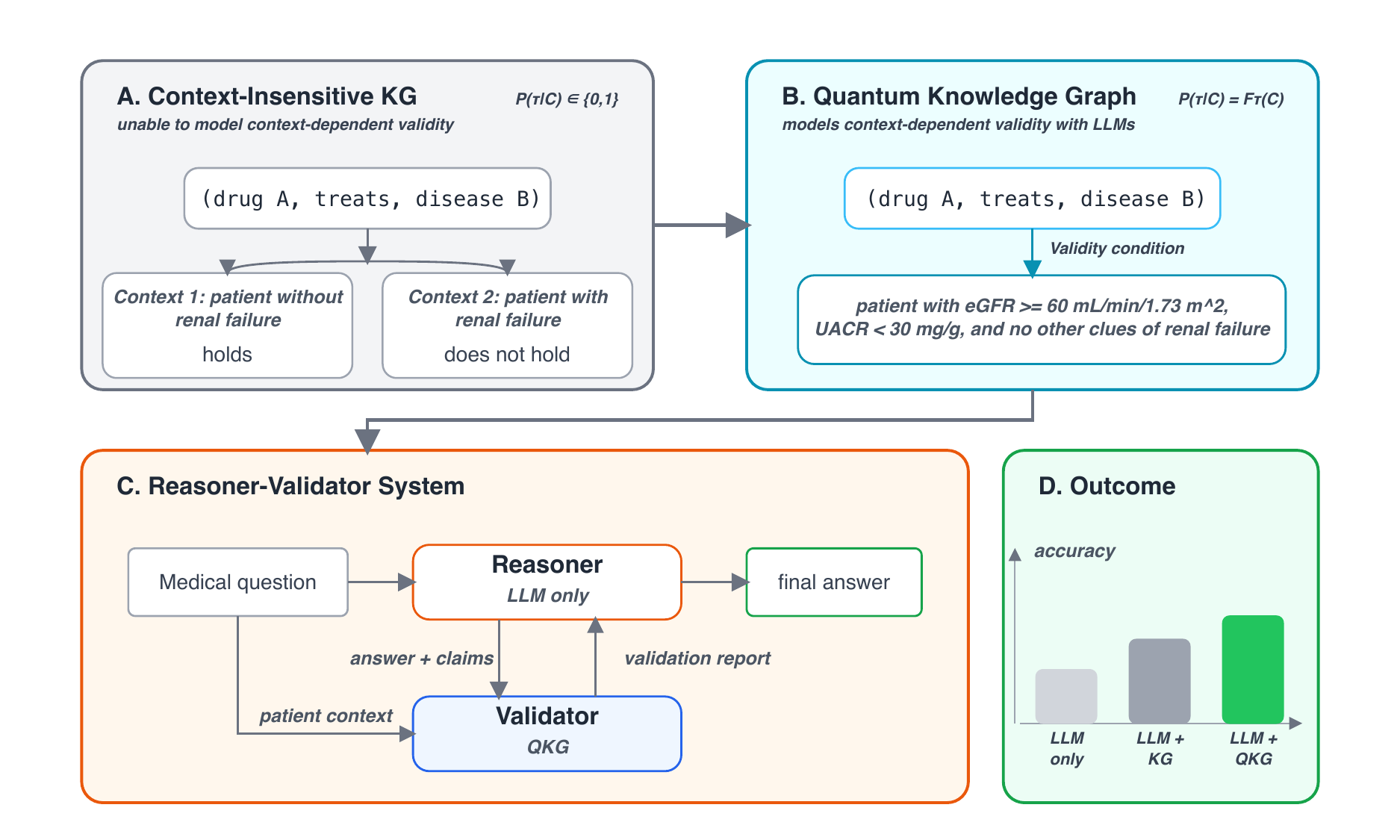}
  \caption{Overview of the proposed Quantum Knowledge Graph (QKG) framework. Panel A illustrates the limitation of conventional knowledge graph triples, whose validity is effectively reduced to a context-insensitive binary assignment and therefore may hold in one patient context but fail in another. Panel B shows how QKG operationalizes the triplet-specific function \(F_{\tau}(C)\) by attaching natural-language validity conditions, enabling context-dependent triplet evaluation with LLMs. Panel C presents the reasoner-validator setup used in this work, where a reasoner generates an answer and its supporting claims, and a QKG-based validator evaluates those claims against the patient context and graph knowledge. Panel D summarizes the qualitative performance trend observed in our experiments, where QKG-based validation outperforms validation with the original KG and the no-validator baseline.}
  \label{fig:qkg-overview}
\end{figure}

% Section 5: contributions
In summary, this paper makes the following contributions: 1) we introduce a triplet-validity framing in which the applicability of a knowledge statement is modeled as a context-dependent quantity \(P(\tau \mid C)\), and operationalize its triplet-specific form \(F_{\tau}(C)\) as a Quantum Knowledge Graph (QKG); 2) we instantiate this formulation in medicine by curating a QKG derived from PrimeKG and building a QKG-based validator agent; and 3) we show that integrating QKG-based validation into an LLM reasoner-validator pipeline improves medical question answering over relevant baselines. All curated QKG data and experimental code will be open-sourced.

\section{Background}

\subsection{Context-Dependent Validity in Knowledge Graphs}

Conventional knowledge graphs represent facts as triples $(h,r,t)$ and usually treat each triple as globally valid once it is stored. A substantial body of work has already shown that this assumption is too restrictive. One direction is to enrich triples with qualifiers or additional attributes. Hyper-relational KG methods, such as StarE~\cite{galkin2020stare}, explicitly model relation-specific qualifiers and demonstrate that many facts are better understood as statements that hold together with auxiliary conditions rather than as isolated triples. Another direction is temporalization: temporal KG question answering and reasoning methods allow facts to hold only during specific intervals or at specific granularities, showing that validity may depend on time rather than being universal~\cite{saxena2021cronkgqa,chen2023multitq,ding2024htkg}. Taken together, these works establish an important general point: triplet validity is often conditional.

At the same time, existing contextual extensions usually operationalize context through a limited and pre-specified structure. Hyper-relational KGs assume that relevant contextual dimensions can be attached as explicit qualifiers, while temporal KGs focus primarily on time. These are important advances, but they do not fully address settings where applicability depends on richer and more compositional conditions that are difficult to enumerate in advance. In such cases, the central problem is no longer whether a fact exists in the graph, but in what context that fact should be regarded as valid.

\subsection{Context-Dependent Validity in Medical Knowledge Graphs}

This limitation is especially visible in medicine, where the validity of a knowledge statement often depends on patient-specific details such as comorbidities, laboratory findings, disease stage, medication history, and contraindications. Biomedical KGs such as PrimeKG~\cite{primekg} provide broad relational coverage for precision medicine, but their edges mainly record that an association exists, not the exact conditions under which it should be applied to a particular patient. Prior medical KG research has already moved beyond the plain-triple formulation in two relevant ways. Li et al.~\cite{li2020realworld} propose a real-world medical KG with a quadruplet structure, showing that clinical facts often require richer factual representation than a bare $(h,r,t)$ tuple. Li et al.~\cite{li2020probkg} further introduce a probabilistic medical KG embedding method that models uncertainty at the triplet level, moving from binary validity toward population-level confidence. These studies are important because they show that medical knowledge is neither purely context-free nor strictly deterministic.

Related work in clinical temporal knowledge graphs reinforces the same point from a different angle. Diao et al.~\cite{diao2021clinicaltkg} model temporal clinical KGs for diabetic complication prediction, showing that medical knowledge use is often inseparable from evolving clinical context. However, richer schemas, temporalization, and probabilistic weighting still do not directly provide a mechanism for deciding whether a specific triplet is applicable to a specific patient in a given question. Benchmarks such as MedReason~\cite{wu2025medreason} make the need for KG-grounded medical reasoning concrete: the task is not only to retrieve medically related entities and relations, but also to organize them into reasoning paths that are consistent with clinical logic and evidence-based medicine. This motivates treating triplet validity itself as context-dependent, rather than assuming that retrieval of relevant facts is sufficient.

\section{Method}

% TODO: If needed, add a short conceptual note here on two possible ways to realize triplet-specific functions F_tau(C):
% (1) enrich relations with explicit applicability conditions; or
% (2) instantiate entities under a specific scenario so evaluation is over context-specific entities.
% The current paper adopts the first route.

\subsection{Knowledge Sources}

\subsubsection{Disease-Centric Subgraph from PrimeKG}
PrimeKG~\cite{primekg} provides the source biomedical knowledge graph for this work. Working with the full graph is computationally prohibitive and introduces noise irrelevant to a given clinical domain. We therefore construct a focused subgraph centered on a target disease entity---in our experiments, \textit{diabetes mellitus} (MONDO:5015). The construction proceeds in two layers. The \textbf{direct layer} collects all triplets $(h, r, t) \in \text{PrimeKG}$ in which either $h$ or $t$ is the target disease entity, yielding the \textit{intermediate entity set} $\mathcal{E}_1$ of entities one hop away from diabetes. The \textbf{indirect layer} then collects all triplets in which at least one endpoint belongs to $\mathcal{E}_1$, capturing second-order associations---e.g., drugs that act on proteins involved in diabetes-related pathways---without expanding to the entire graph. The two layers are merged and deduplicated to form the final subgraph $\mathcal{G}_{\text{sub}}$. The direct layer yields 1{,}470 triplets and $|\mathcal{E}_1|=735$ intermediate entities; the indirect layer contributes a further 861{,}070 triplets. After deduplication, $\mathcal{G}_{\text{sub}}$ contains \textbf{862{,}540 triplets} across \textbf{18{,}387 unique entities} spanning 10 biomedical entity types (gene/protein, drug, disease, biological process, phenotype, pathway, exposure, molecular function, cellular component, and anatomy) and 25 relation types.

\subsubsection{Focused Relation Annotation}
Most PrimeKG relation types encode biological or molecular facts whose validity is relatively stable across patient contexts. We therefore focus on relation types whose applicability is more likely to vary with patient-specific factors: \texttt{indication}, \texttt{contraindication}, \texttt{off-label use}, and \texttt{drug\_effect}. For each unique triplet $(h, r, t)$ over these four types, we use the Baichuan-M2-Plus API~\cite{baichuanM2blog} to generate evidence about population-specific applicability. The outputs are stored as structured \texttt{ConstraintItem} records, each containing the \texttt{patient\_characteristics} in which the relation holds (e.g., ``eGFR $<$ 30'', ``HbA1c $>$ 9\%''), an \texttt{applicability} level drawn from a five-point ordinal scale (\textit{Definitely Applicable} through \textit{Definitely NOT Applicable}), and supporting \texttt{evidence} text. The resulting \texttt{relation\_with\_facts} collection contains 68{,}651 annotated facts spanning 2{,}591 unique entities and 4 relation types. These annotations are retrieved at inference time to support patient-context filtering.

\subsection{Reasoning Pipeline}
We implement a two-agent loop in which a pure-LLM \textbf{Reasoner} and a KG-grounded \textbf{Validator} collaborate iteratively (Figure~\ref{fig:qkg-overview}, Panel C). The Reasoner first proposes an answer and emits structured claims; the Validator then checks each claim against the QKG and the patient context; finally, the Reasoner reconsiders its answer in light of the resulting validation report. To perform this validation step, the patient context is derived directly from the clinical question, including demographic factors, comorbidities, laboratory values, and current medications. For each retrieved KG relation, the Validator examines its associated \texttt{ConstraintItem} records and determines whether the constraint applies to the current patient. Relations whose constraints are not met are down-weighted or excluded before being used as evidence, allowing claim verification to be conditioned on the patient context rather than on raw graph connectivity alone. In our implementation, the validator is allowed up to 20 tool-use turns per round. Algorithm~\ref{alg:pipeline} summarizes the full procedure.

\subsection{Statistical Testing}
\label{sec:statistical-testing}
To test whether accuracy differences between two pipeline settings are statistically significant, we apply McNemar's test on paired per-sample correctness. For settings $A$ and $B$ evaluated on the same questions, let $b$ and $c$ denote the counts of samples that are correct under $A$ but wrong under $B$, and wrong under $A$ but correct under $B$, respectively. Under the null hypothesis that each discordant flip is equally likely in either direction, we report the exact two-sided binomial $p$-value $p = \min\{1,\, 2 \sum_{k=\max(b,c)}^{b+c} \binom{b+c}{k} 2^{-(b+c)}\}$. For comparisons against the no-validator baseline, the reasoner-only correctness is treated as condition $A$ and the validated final correctness as condition $B$ on the same run. For leakage-adjusted comparisons on the Qwen-validator runs, samples are removed before the paired test if, in either run, their W$\to$C revision was labelled \texttt{LIKELY\_LEAKAGE} or their C$\to$W regression was labelled \texttt{LIKELY\_KG\_SUPPORTED} with decisive evidence citing a QKG applicability token (the ctx-driven subset). This matches the per-run adjustment in Eq.~\ref{eq:adj-final-acc} (Appendix~\ref{app:leakage-rules}).

\begin{algorithm}[t]
\caption{QKG Reasoning Pipeline}
\label{alg:pipeline}
\begin{algorithmic}[1]
\Require question $Q$, choices $C$, knowledge graph $\mathcal{G}_{\text{sub}}$
\Ensure final answer $A^*$

\State $P \leftarrow \textsc{ExtractPatientContext}(Q)$
\State $(A,\, \text{claims}) \leftarrow \textsc{Reasoner}(Q, C)$
\Comment{pure LLM; emits claims for answer options}

\For{each claim $c \in \text{claims}$}
    \State $\mathcal{E} \leftarrow \textsc{SearchEntities}(c)$
    \State $\mathcal{R} \leftarrow \textsc{GetRelationsWithContext}(\mathcal{E},\, \mathcal{G}_{\text{sub}})$
    \State $\mathcal{R}_P \leftarrow \textsc{ApplyConstraintItems}(\mathcal{R}, P)$
    \If{$\mathcal{R}_P$ supports $c$}
        \State $\text{status}(c) \leftarrow \texttt{SUPPORTED}$
    \ElsIf{$\mathcal{R}_P$ contradicts $c$}
        \State $\text{status}(c) \leftarrow \texttt{CONTRADICTED}$
    \Else
        \State $\text{status}(c) \leftarrow \texttt{NO\_COVERAGE}$
    \EndIf
\EndFor

\State $\text{report} \leftarrow \{(c,\, \text{status}(c)) \mid c \in \text{claims}\}$

\If{any claim in $\text{report}$ is \texttt{CONTRADICTED}}
    \State $A^* \leftarrow \textsc{Reasoner}(Q, C, \text{report})$ \Comment{reconsider}
\Else
    \State $A^* \leftarrow A$
\EndIf
\State \Return $A^*$
\end{algorithmic}
\end{algorithm}

\section{Experimental Setup}

This section describes the datasets, evaluation protocol, and compared settings used in our experiments.

\subsection{Datasets}

\label{sec:dataset}
We evaluate our approach on medical question answering using samples from MedReason~\cite{wu2025medreason}, a medical reasoning dataset of approximately 30{,}000 questions built from seven source QA datasets: MedQA, MedMCQA, PubMedQA, MMLU, MedXpert, HuatuoGPT-o1, and the medical subset of Humanity's Last Exam (HLE). To ensure alignment with our curated PrimeKG subset, we construct a KG-grounded evaluation set by extracting question and option entities, grounding them to UMLS, aligning them to PrimeKG nodes, filtering out samples with no recoverable KG paths, and annotating patient characteristics from the question text for downstream context matching. The resulting evaluation set contains \textbf{2{,}788 samples} spanning a range of diabetes-related clinical scenarios with verified KG coverage. Detailed dataset construction steps and a table reporting QA-source distributions are provided in Appendix~\ref{app:dataset-construction}.

\subsection{Evaluation Protocol}

For each sample, the model generates a single answer (A--J) along with structured reasoning. Outputs are constrained using a Pydantic schema (\texttt{QAResponse}; see Appendix~\ref{app:qaresponse-schema}) to ensure consistent JSON formatting. Evaluation is fully automated by parsing model outputs and computing exact-match accuracy against the gold answers.

The primary evaluation metric is exact-match accuracy, defined as the proportion of predictions that match the gold answer. We also report secondary metrics for the reasoner-validator pipeline, including the number and percentage of cases whose answers change after validation, as well as how often those revisions improve or degrade final correctness.

\subsection{Models and Compared Settings}

The main experiments use two LLMs: \textbf{Haiku-4.5}~\cite{anthropic2025haiku45} and \textbf{Qwen-3.6-Plus}~\cite{qwen2026qwen36plus}. We treat Haiku-4.5 as the baseline model and Qwen-3.6-Plus as the higher-capability model in our study setup. We compare three settings: a no-validator baseline, KG validation without context matching, and QKG validation with context matching. These settings are used to study both patient-context ablation and model-capacity effects in the reasoner--validator pipeline.

\section{Results}

\subsection{Main Results and Patient-Context Ablation}

\begin{figure}[t]
\centering
\includegraphics[width=\linewidth]{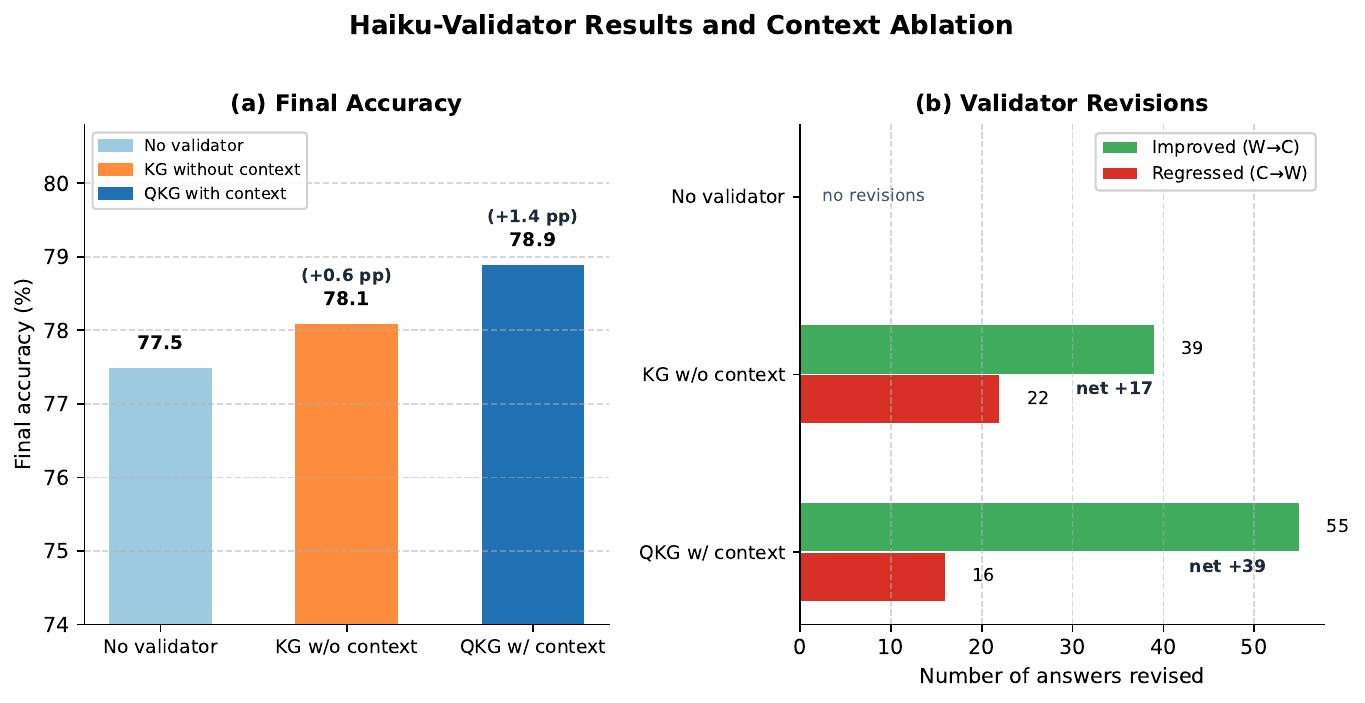}
\caption{Haiku-validator results and context ablation on the curated evaluation set ($N=2{,}788$), using Haiku-4.5 as the Reasoner throughout. Panel (a) shows final accuracy for the no-validator baseline, KG validation without context matching, and QKG validation with context matching; the two validation settings use Haiku-4.5 as the Validator. Panel (b) shows the number of answers revised by validation, separated into wrong-to-correct improvements and correct-to-wrong regressions. Paired McNemar tests (exact two-sided; Section~\ref{sec:statistical-testing}) confirm all three pairwise differences: baseline vs.\ KG w/o context $p{=}0.04$, baseline vs.\ QKG w/ context $p{\approx}3.8{\times}10^{-6}$, and KG w/o context vs.\ QKG w/ context $p{=}0.014$.}
\label{fig:main_patient_results}
\end{figure}

Figure~\ref{fig:main_patient_results} shows the main results on the curated evaluation set ($N=2{,}788$). Across all three settings, Haiku-4.5 is used as the Reasoner; in the two validation settings, Haiku-4.5 is also used as the Validator. The main ablation compares KG validation without context matching against QKG validation with context matching.

Validation changes 2.19\% of answers in the no-context setting (39 W$\to$C, 22 C$\to$W) and 2.55\% in the with-context setting (55 W$\to$C, 16 C$\to$W), so the with-context setting also yields a higher wrong-to-correct rate and a lower correct-to-wrong rate.

\subsection{Case Studies of Context-Dependent Correction}

\begin{figure}[t]
\centering
\includegraphics[width=\linewidth]{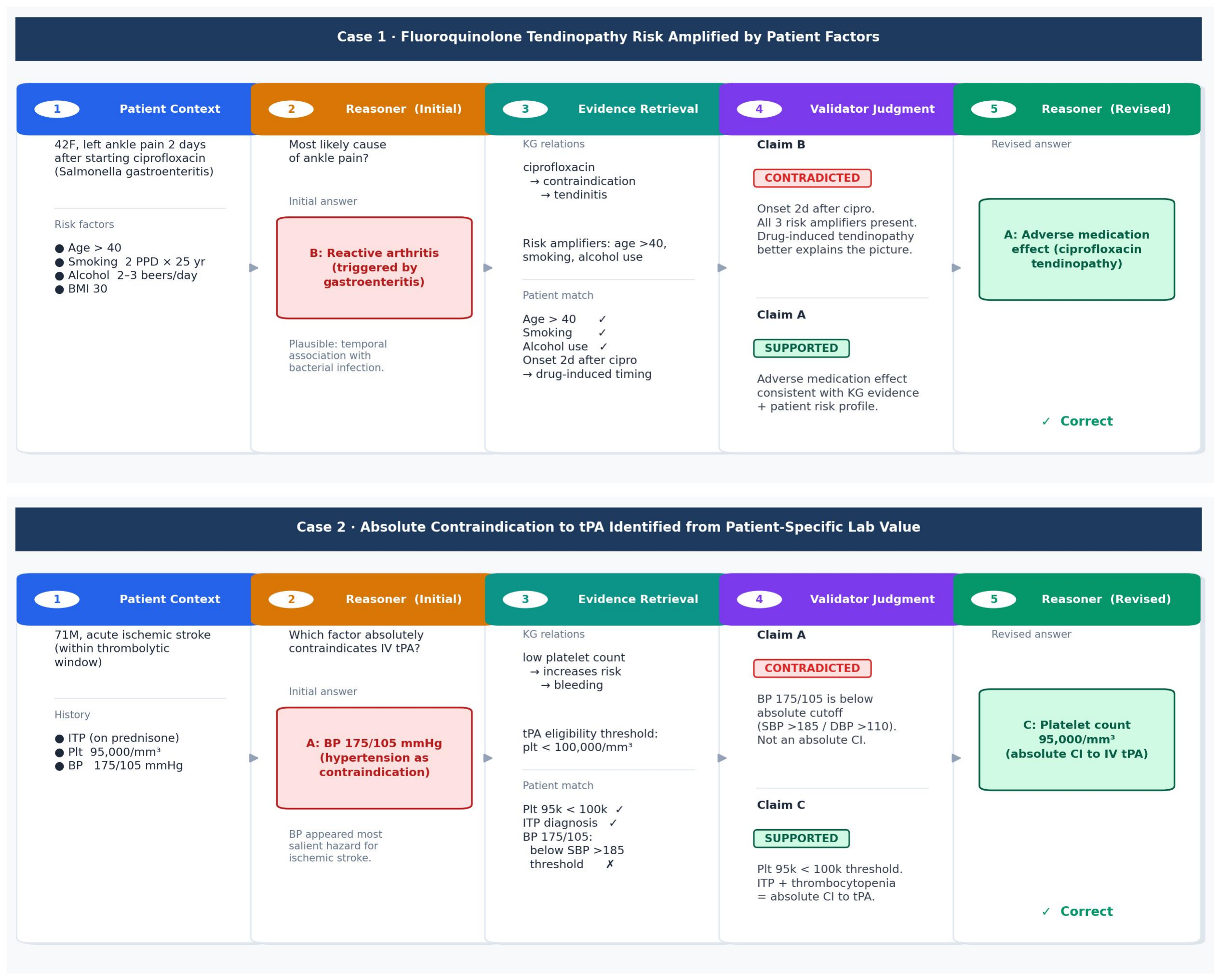}
\caption{Two case studies of context-dependent correction. The top panel shows a compositional risk-amplifier case, where the validator combines multiple patient-specific factors to identify fluoroquinolone-associated tendinopathy. The bottom panel shows a threshold-based contraindication case, where the validator matches a patient-specific platelet count to the eligibility threshold for IV tPA.}
\label{fig:case-studies}
\end{figure}

Figure~\ref{fig:case-studies} presents two case studies of context-dependent correction. The top case is a compositional patient-context example, where the validator combines multiple patient-specific factors---age, smoking, alcohol use, and temporal proximity to ciprofloxacin exposure---to revise the initial answer. The bottom case is a threshold-based example, where the validator matches a patient-specific platelet count of 95{,}000/mm$^3$ against the eligibility threshold for IV tPA.

\subsection{Qwen-3.6-Plus as Validator}

\begin{figure}[t]
\centering
\includegraphics[width=\linewidth]{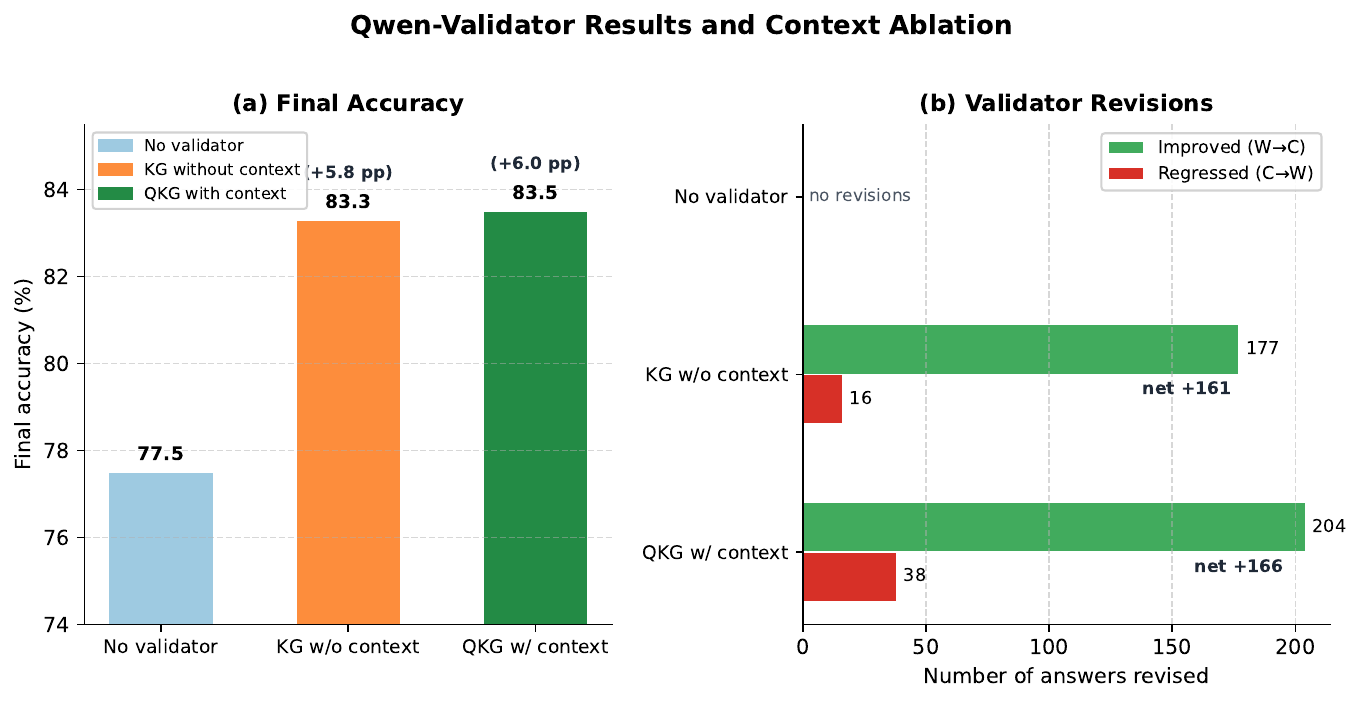}
\caption{Qwen-validator results and context ablation on the curated evaluation set ($N=2{,}788$), with Haiku-4.5 used as the Reasoner throughout. Panel (a) shows final accuracy for the no-validator baseline, KG validation without context matching, and QKG validation with context matching. Panel (b) shows the corresponding wrong-to-correct improvements and correct-to-wrong regressions.}
\label{fig:qwen-validator-results}
\end{figure}

Figure~\ref{fig:qwen-validator-results} shows the strong-validator comparison on the same curated evaluation set ($N=2{,}788$). Across all three settings, Haiku-4.5 is used as the Reasoner. The comparison then varies whether validation is absent, performed with KG evidence without context matching, or performed with QKG validation with context matching, using Qwen-3.6-Plus as the Validator in the latter two settings. Qwen-3.6-Plus is substantially stronger than Haiku-4.5 on this set---its standalone accuracy as a reasoner is 89.1\%, against the 77.5\% Haiku-4.5 reasoner baseline---so this configuration pairs a weaker Reasoner with a stronger Validator, which is the setting in which validator-supplied prior knowledge is most likely to contaminate the measured validation gain.

\paragraph{Case studies of strong-validator answer leakage.}
Two W$\to$C revisions from the no-context Qwen-3.6-Plus validator run illustrate how leakage manifests. In qa\_9542 (gold: shingles vaccine), the KG provides no directly relevant scheduling edge for any option, yet the elimination of the gold answer is still flagged \texttt{CONTRADICTED} on a seasonal-influenza-timing argument the Validator supplies itself. In qa\_6324 (gold: antibiotic prophylaxis before molar extraction), two \texttt{CONTRADICTED} statuses similarly rest on validator-supplied medical knowledge after the KG lookup fails to return a directly relevant edge---one citing AHA prophylaxis guidance for the gold answer, the other citing general pharmacology of nitrous oxide and trapped gas spaces for the eliminated option. Full vignettes, quoted evidence, and the with-context status pattern on these samples are in Appendix~\ref{app:leakage-cases}.

\paragraph{Quantitative leakage classification.}
To estimate how often validator answer leakage drives W$\to$C revisions, we label each W$\to$C case in both Qwen-validator runs as \emph{likely KG-supported}, \emph{mixed}, or \emph{likely leakage}, based on whether the validator's decisive \texttt{CONTRADICTED} evidence cites a KG entity/relation (or a QKG applicability annotation), pivots from a KG gap to clinical-guideline knowledge, or sits between the two. The exact rules, the LLM re-labelling step for cases the regex leaves indeterminate, the leakage-adjusted accuracy formula, and the released per-case CSV are described in Appendix~\ref{app:leakage-rules}.

Table~\ref{tab:leakage-classification} summarizes the high-level accounting, and Table~\ref{tab:classification-breakdown} reports the class-level breakdown of W$\to$C and C$\to$W revisions. Both runs leak comparably ($\sim$55--60 W$\to$C revisions labeled likely leakage out of $\sim$200; Table~\ref{tab:leakage-classification}), but the with-context run produces noticeably more KG-supported W$\to$C revisions (123 vs.\ 97; Table~\ref{tab:classification-breakdown}), and 27 of those explicitly invoke a QKG-specific applicability token (\texttt{AVOID}, \texttt{RECOMMENDED}, \texttt{CAUTION}, \texttt{ConstraintItem}, or \texttt{safety judgment}) versus 0 in the no-context run, indicating that the patient-context-conditioned QKG mechanism is responsible for that excess. After dropping the likely-leakage W$\to$C revisions and the ctx-driven KG-supported C$\to$W regressions from both numerator and denominator (Appendix~\ref{app:leakage-rules}, Eq.~\ref{eq:adj-final-acc}), the leakage-adjusted final accuracies are 82.88\% in the no-context setting and 83.75\% in the with-context setting; both still exceed the no-validator Reasoner baseline (77.5\%) by more than 5 percentage points, and the with-context setting still exceeds the no-context setting after adjustment.

\begin{table}[t]
\centering
\caption{Leakage-adjusted accuracy accounting for the two Qwen-3.6-Plus validator runs (Haiku-4.5 Reasoner throughout). W$\to$C and C$\to$W are the counts of wrong-to-correct revisions and correct-to-wrong regressions after validation. Est.\ adj.\ is the total number of samples dropped from the Adj.\ final accuracy; exclusion rules and the formula are in Appendix~\ref{app:leakage-rules}, Eq.~\ref{eq:adj-final-acc}. Class-level breakdowns of W$\to$C and C$\to$W are reported in Table~\ref{tab:classification-breakdown}. A paired McNemar test for the with-context vs.\ no-context comparison gives exact two-sided $p{=}0.73$ on the raw paired set ($N{=}2{,}782$, $b{=}65$, $c{=}70$), and $p{=}0.05$ on the leakage-adjusted subset ($N{=}2{,}665$, $b{=}33$, $c{=}52$; samples flagged in either run removed). See Section~\ref{sec:statistical-testing}.}
\label{tab:leakage-classification}
\small
\setlength{\tabcolsep}{6pt}
\begin{tabular}{lccccccc}
\toprule
 &  &  &  &  & \multicolumn{2}{c}{Final Acc.\ (\%)} \\
\cmidrule(lr){6-7}
Setting & $N$ & W$\to$C & C$\to$W & Est.\ adj. & Raw & Adj. \\
\midrule
KG w/o context  & 2{,}788 & 177 & 16 & 60 & 83.25 & 82.88 \\
QKG w/ context  & 2{,}788 & 204 & 38 & 75 & 83.46 & 83.75 \\
\bottomrule
\end{tabular}
\end{table}

\paragraph{Leakage classification of C$\to$W regressions.}
Applying the same classifier to the C$\to$W regressions in both Qwen-validator runs tests whether the elevated C$\to$W under QKG w/ context reflects correct patient-context-conditioned elimination of an option whose underlying fact is MCQ-gold (KG-supported), or validator-supplied prior knowledge that misled the Reasoner away from gold (leakage). A C$\to$W is decisive when a \texttt{CONTRADICTED} claim either contradicts the option the Reasoner originally chose (the gold) or un-eliminates the option that eventually became the final (wrong) answer; the regex rules and the LLM re-labeling pass on \textsc{Unclassified} cases are the same as for the W$\to$C classification. The resulting per-class counts are reported alongside the W$\to$C breakdown in Table~\ref{tab:classification-breakdown}. Of the 38 C$\to$W in the with-context run, 36 are KG-supported and only 1 is likely leakage; 20 of the 36 KG-supported cases explicitly cite a QKG applicability token, versus 0 of the 12 KG-supported cases in the no-context run. The increase from 16 to 38 C$\to$W under QKG is therefore dominated by KG-supported regressions (+24) and by QKG-token-driven regressions specifically (+20), not by validator hallucination.

\begin{table}[t]
\centering
\caption{Class-level leakage-classification breakdown of W$\to$C (wrong-to-correct) revisions and C$\to$W (correct-to-wrong) regressions in the two Qwen-3.6-Plus validator runs, using the rules in Appendix~\ref{app:leakage-rules}. The parenthesised count under KG-supp.\ is the subset whose decisive evidence cites a QKG-specific applicability token (\texttt{AVOID}/\texttt{RECOMMENDED}/\texttt{CAUTION}/\texttt{ConstraintItem}/\texttt{safety judgment}). Each block's class columns sum to the W$\to$C or C$\to$W totals in Table~\ref{tab:leakage-classification}.}
\label{tab:classification-breakdown}
\small
\setlength{\tabcolsep}{6pt}
\begin{tabular}{lccccccc}
\toprule
 & \multicolumn{3}{c}{W$\to$C class counts} &  & \multicolumn{3}{c}{C$\to$W class counts} \\
\cmidrule(lr){2-4} \cmidrule(lr){6-8}
Setting & KG-supp.\ (ctx) & Mixed & Leakage & & KG-supp.\ (ctx) & Mixed & Leakage \\
\midrule
KG w/o context  & 97 (0)   & 20 & 60 & & 12 (0)  & 0 & 4 \\
QKG w/ context  & 123 (27) & 26 & 55 & & 36 (20) & 1 & 1 \\
\bottomrule
\end{tabular}
\end{table}

\section{Discussion}

\subsection{Not Merely Facts, but Applicable Knowledge}

In clinical reasoning, the key issue is not whether a KG fact is medically related, but whether that fact is applicable in the patient's specific context. This is the central distinction highlighted by the case studies in Section~5.2. In the first case, the validator succeeds only after combining multiple patient-specific factors---age, smoking, alcohol use, and timing after ciprofloxacin exposure---to determine that the drug-induced tendinopathy relation is applicable to this particular patient. In the second case, the validator succeeds only by matching a concrete laboratory value, a platelet count of 95{,}000/mm$^3$, against the threshold for tPA eligibility. One case is compositional and multifactorial; the other is threshold-based. Together, they show that the relevant clinical knowledge cannot be treated as context-free once retrieved from the graph.

This is exactly the motivation for modeling triplet validity under context. If a KG edge is treated as simply true once retrieved, the validator cannot distinguish between knowledge that is generally relevant and knowledge that is actually valid for the current patient. In other words, the failure mode is not simply missing facts; it is the inability to decide when a retrieved fact should count as evidence. QKG addresses this problem by attaching conditions under which a triplet should be accepted, contradicted, or ignored in the current patient context.

The aggregate results are consistent with this interpretation. Figure~\ref{fig:main_patient_results} suggests that merely adding KG-backed knowledge is already somewhat useful, since KG validation without context matching still improves over the no-validator baseline. But the larger gain comes from context-aware validation: the full QKG setting performs better because it does not stop at retrieving medically related knowledge, and instead decides whether a retrieved relation should count for this patient. Figure~\ref{fig:qwen-validator-results} preserves the same ordering under a stronger validator, with QKG with context matching still outperforming KG validation without context matching, which in turn outperforms the no-validator baseline. Though the Qwen-as-validator result makes the gap between with-context and without-context settings appear relatively small, in the next subsection we argue that this pattern is compounded by the strong validator's own model-internal knowledge. Even with that caution, the repeated ordering still supports the central idea that applicability carries useful signal beyond raw factual relatedness alone.

\subsection{Why Strong-Validator Results Require Caution}

The Qwen-3.6-Plus validator results are harder to interpret causally than the Haiku-validator ablation. The case studies in Section~5.3 and the log analysis summarized in Table~\ref{tab:leakage-classification} show that both Qwen-validator runs contain answer leakage: some wrong-to-correct revisions are driven by validator-supplied medical or guideline knowledge rather than by a directly relevant retrieved edge, so the raw gains in both Qwen settings are not clean measurements of graph-grounded validation.

We read both the paired-test null ($p{=}0.73$) between raw accuracy w/ context and w/o context and the elevated C$\to$W under w/ context (38 vs.\ 16) in Table~\ref{tab:leakage-classification} as symptoms of the benchmark's fact-level rather than patient-level gold, not of a noisy QKG mechanism. Applying the per-case leakage classifier to the C$\to$W regressions (Table~\ref{tab:classification-breakdown}) supports this reading directly: 36 of 38 QKG-w/-context C$\to$W cases are KG-supported and only 1 is likely leakage, with 20 of the KG-supported cases explicitly citing QKG applicability tokens (versus 0 of 12 KG-supported C$\to$W in the no-context run). The elevated C$\to$W under QKG is therefore dominated by patient-context-conditioned reasoning that correctly eliminates an option whose underlying fact is nonetheless the one the MCQ gold rewards, not by validator-supplied prior knowledge at the applicability step. Removing these benchmark-gold-noise cases from the paired comparison confirms the reading quantitatively: the leakage-adjusted paired McNemar (Section~\ref{sec:statistical-testing}) shifts the with-context vs.\ no-context $p$-value from $0.73$ on the raw set to $0.05$ on the adjusted set---right at the conventional $\alpha{=}0.05$ threshold, turning a strong null into a borderline-significant effect once the benchmark-gold-noise cases are removed.

QKG's context effect should be most material in real-world clinical reasoning, where answers routinely depend on context-conditioned combinations of evidence rather than single-fact recall. In the absence of a suitable real-world benchmark (see Limitation and Future Work below), we evaluate on MCQ medical QA, whose fact-level gold cannot fully expose this regime. On this benchmark, Figure~\ref{fig:main_patient_results}---the Haiku-validator context effect of $+0.79$ pp with paired McNemar $p{=}0.014$---is the cleanest observation of the mechanism, whereas the Qwen-validator picture requires the leakage adjustment to unmask the effect ($p$ from $0.73$ to $0.05$) against the benchmark's fact-level gold.

\subsection{Limitations and Future Work}

The main limitation of the current study is that evaluation on benchmark medical QA cannot fully disentangle QKG-based contextual validation from model-internal medical knowledge. This issue becomes especially important for strong validators, whose gains may reflect both explicit use of patient-conditioned KG evidence and prior familiarity with benchmark-style medical questions. A cleaner test would use real-world patient-level reasoning tasks, but such evaluation remains difficult because publicly available clinical datasets rarely provide scalable gold-standard labels for contextual reasoning, while routine clinical data are often noisy, incomplete, and not annotated with unambiguous reasoning traces. Existing open clinical resources are highly valuable, but they are usually optimized either for general EHR access, as in MIMIC-IV~\cite{johnson2023mimiciv}, or for predictive benchmarking, as in EHRSHOT's few-shot patient classification tasks~\cite{wornow2023ehrshot}, rather than for evaluating whether a system can determine which knowledge is applicable under a patient's specific context. The current results should therefore be interpreted as strong benchmark evidence for the usefulness of QKG, but not yet as a definitive causal measurement of its contribution in real-world clinical workflows.

In future work, we aim to build such a real-world clinical reasoning benchmark and share it with the community. Our goal is to use it not only to test QKG under more realistic clinical conditions, but also to diagnose its failure modes and improve context-dependent KG validation at larger scale.

\section{Conclusion}

This paper introduces the Quantum Knowledge Graph (QKG), a framework for modeling triplet validity as context-dependent rather than context-insensitive. We instantiate QKG in the medical domain by augmenting KG relations with natural-language applicability conditions and using them in a reasoner--validator pipeline for medical question answering. Under a matched Haiku-4.5 Reasoner--Validator setting, patient-context matching delivers a small but paired-significant gain over KG validation without context ($+0.79$ pp, $p{=}0.014$), and both settings exceed the no-validator baseline. Under a stronger validator (Qwen-3.6-Plus), the raw paired gap is a null ($p{=}0.73$) that becomes borderline significant ($p{=}0.05$) after adjusting for knowledge leakage and suspicious questions---consistent with a benchmark-gold ceiling in multiple-choice medical QA rather than QKG redundancy---and we propose real-world clinical reasoning tasks as a direct next step for testing context-dependent KG validation at larger scale. More broadly, the findings support the view that the value of a knowledge graph in LLM-based reasoning lies not only in storing relevant facts, but in representing whether those facts are applicable in the specific context in which they are used.

\section*{Acknowledgements}
This work was supported by City University of Hong Kong under project number 9610777. We gratefully acknowledge Baichuan Intelligence for providing complimentary token credits to support our use of the Baichuan M2 Plus model. We also thank Dr. Linfeng Li, lead author of \citet{li2020realworld,li2020probkg}, for his insightful discussions with us, which helped inspire this work. In addition, we thank Prof. Wei-Ying Ma, Director of HKAI-Sci, for encouraging our exploration of this research frontier.

\paragraph{Reproducibility Checklist.}
See Appendix~\ref{app:code-and-data} for release and reproducibility details.

\bibliographystyle{plainnat}
\bibliography{references}

@article{primekg,
  title={PrimeKG: A Knowledge Graph for Precision Medicine},
  author={Chandak, P. and others},
  journal={Scientific Data},
  year={2023}
}

@article{johnson2023mimiciv,
  title={MIMIC-IV, a Freely Accessible Electronic Health Record Dataset},
  author={Johnson, Alistair E. W. and Bulgarelli, Lucas and Shen, Lu and Gayles, Alvin and Shammout, Ayad and Horng, Steven and Pollard, Tom J. and Hao, Sicheng and Moody, Benjamin and Gow, Brian and Lehman, Li-wei H. and Celi, Leo A. and Mark, Roger G.},
  journal={Scientific Data},
  volume={10},
  number={1},
  pages={1},
  year={2023},
  doi={10.1038/s41597-022-01899-x}
}

@inproceedings{wornow2023ehrshot,
  title={EHRSHOT: An EHR Benchmark for Few-Shot Evaluation of Foundation Models},
  author={Wornow, Michael and Thapa, Rahul and Steinberg, Ethan and Fries, Jason A. and Shah, Nigam H.},
  booktitle={Advances in Neural Information Processing Systems 36: Datasets and Benchmarks Track},
  year={2023},
  url={https://proceedings.neurips.cc/paper_files/paper/2023/file/d42db1f74df54cb992b3956eb7f15a6f-Paper-Datasets_and_Benchmarks.pdf}
}

@article{li2020realworld,
  title={Real-world data medical knowledge graph: construction and applications},
  author={Li, Linfeng and Wang, Peng and Yan, Jun and Wang, Yao and Li, Simin and Jiang, Jinpeng and Sun, Zhe and Tang, Buzhou and Chang, Tsung-Hui and Wang, Shenghui and Liu, Yuting},
  journal={Artificial Intelligence in Medicine},
  volume={103},
  pages={101817},
  year={2020},
  doi={10.1016/j.artmed.2020.101817}
}

@article{li2020probkg,
  title={A Method to Learn Embedding of a Probabilistic Medical Knowledge Graph: Algorithm Development},
  author={Li, Linfeng and Wang, Peng and Wang, Yao and Wang, Shenghui and Yan, Jun and Jiang, Jinpeng and Tang, Buzhou and Wang, Chengliang and Liu, Yuting},
  journal={JMIR Medical Informatics},
  volume={8},
  number={5},
  pages={e17645},
  year={2020},
  doi={10.2196/17645},
  pmid={32436854},
  pmcid={7273238}
}

@article{wu2025medreason,
  title={Medreason: Eliciting factual medical reasoning steps in llms via knowledge graphs},
  author={Wu, Juncheng and Deng, Wenlong and Li, Xingxuan and Liu, Sheng and Mi, Taomian and Peng, Yifan and Xu, Ziyang and Liu, Yi and Cho, Hyunjin and Choi, Chang-In and others},
  journal={arXiv preprint arXiv:2504.00993},
  year={2025}
}

@article{diao2021clinicaltkg,
  title={The research of clinical temporal knowledge graph based on deep learning},
  author={Diao, Lijuan and Yang, Wei and Zhu, Penghua and Cao, Gaofang and Song, Shoujun and Kong, Yang},
  journal={Journal of Intelligent \& Fuzzy Systems},
  volume={41},
  number={3},
  pages={4265--4274},
  year={2021},
  doi={10.3233/JIFS-189687}
}

@misc{baichuanM2blog,
  author={{Baichuan Intelligence}},
  title={Baichuan-M2 Technical Blog},
  howpublished={\url{https://www.baichuan-ai.com/blog/baichuan-M2}},
  year={2025},
  note={Accessed 2026-04-13}
}

@misc{anthropic2025haiku45,
  author={{Anthropic}},
  title={Introducing Claude Haiku 4.5},
  year={2025},
  month={10},
  howpublished={\url{https://www.anthropic.com/news/claude-haiku-4-5}},
  note={Release note; accessed 2026-04-17}
}

@misc{qwen2026qwen36plus,
  author={{Qwen Team}},
  title={Qwen3.6-Plus: Towards Real World Agents},
  year={2026},
  month={4},
  howpublished={\url{https://qwen.ai/blog?email_hash=0d7a7050906b225db2718485ca0f3472&id=qwen3.6}},
  note={Release note; accessed 2026-04-17}
}

@inproceedings{dougrezlewis2025recv,
  title={Assessing the Reasoning Capabilities of LLMs in the context of Evidence-based Claim Verification},
  author={Dougrez-Lewis, John and Akhter, Mahmud Elahi and Ruggeri, Federico and L{\"o}bbers, Sebastian and He, Yulan and Liakata, Maria},
  booktitle={Findings of the Association for Computational Linguistics: ACL 2025},
  pages={20604--20628},
  year={2025},
  publisher={Association for Computational Linguistics},
  url={https://aclanthology.org/2025.findings-acl.1059/},
  doi={10.18653/v1/2025.findings-acl.1059}
}

@inproceedings{kolli2025hybrid,
  title={Hybrid Fact-Checking that Integrates Knowledge Graphs, Large Language Models, and Search-Based Retrieval Agents Improves Interpretable Claim Verification},
  author={Kolli, Shaghayegh and Rosenbaum, Richard and Cavelius, Timo and Strothe, Lasse and Lata, Andrii and Diesner, Jana},
  booktitle={Proceedings of the 9th Widening NLP Workshop},
  pages={106--115},
  year={2025},
  publisher={Association for Computational Linguistics},
  url={https://aclanthology.org/2025.winlp-main.19/},
  doi={10.18653/v1/2025.winlp-main.19}
}

@inproceedings{galkin2020stare,
  title={Message Passing for Hyper-Relational Knowledge Graphs},
  author={Galkin, Mikhail and Trivedi, Priyansh and Maheshwari, Gaurav and Usbeck, Ricardo and Lehmann, Jens},
  booktitle={Proceedings of the 2020 Conference on Empirical Methods in Natural Language Processing (EMNLP)},
  pages={7346--7366},
  year={2020},
  publisher={Association for Computational Linguistics},
  url={https://aclanthology.org/2020.emnlp-main.596/},
  doi={10.18653/v1/2020.emnlp-main.596}
}

@inproceedings{saxena2021cronkgqa,
  title={Question Answering over Temporal Knowledge Graphs},
  author={Saxena, Apoorv and Tripathi, Aakash and Talukdar, Partha},
  booktitle={Proceedings of the 59th Annual Meeting of the Association for Computational Linguistics and the 11th International Joint Conference on Natural Language Processing (Volume 1: Long Papers)},
  pages={6663--6676},
  year={2021},
  publisher={Association for Computational Linguistics},
  url={https://aclanthology.org/2021.acl-long.520/},
  doi={10.18653/v1/2021.acl-long.520}
}

@inproceedings{ding2024htkg,
  title={Temporal Fact Reasoning over Hyper-Relational Knowledge Graphs},
  author={Ding, Zejin and Wang, Ning and Liu, Shujian and Zhou, Guodong},
  booktitle={Findings of the Association for Computational Linguistics: EMNLP 2024},
  pages={345--357},
  year={2024},
  publisher={Association for Computational Linguistics},
  url={https://aclanthology.org/2024.findings-emnlp.20/},
  doi={10.18653/v1/2024.findings-emnlp.20}
}

@inproceedings{chen2023multitq,
  title={Multi-granularity Temporal Question Answering over Knowledge Graphs},
  author={Chen, Ziyang and Liao, Jinzhi and Zhao, Xiang},
  booktitle={Proceedings of the 61st Annual Meeting of the Association for Computational Linguistics (Volume 1: Long Papers)},
  pages={11380--11395},
  year={2023},
  publisher={Association for Computational Linguistics},
  url={https://aclanthology.org/2023.acl-long.637/},
  doi={10.18653/v1/2023.acl-long.637}
}

@inproceedings{sui2025trustworthykgllm,
  title={Can Knowledge Graphs Make Large Language Models More Trustworthy? An Empirical Study Over Open-ended Question Answering},
  author={Sui, Yuan and He, Yufei and Ding, Zifeng and Hooi, Bryan},
  booktitle={Proceedings of the 63rd Annual Meeting of the Association for Computational Linguistics (Volume 1: Long Papers)},
  pages={12685--12701},
  year={2025},
  address={Vienna, Austria},
  publisher={Association for Computational Linguistics},
  url={https://aclanthology.org/2025.acl-long.622/},
  doi={10.18653/v1/2025.acl-long.622}
}

@inproceedings{wu2025medgraphrag,
  title={Medical Graph RAG: Evidence-based Medical Large Language Model via Graph Retrieval-Augmented Generation},
  author={Wu, Junde and Zhu, Jiayuan and Qi, Yunli and Chen, Jingkun and Xu, Min and Menolascina, Filippo and Jin, Yueming and Grau, Vicente},
  booktitle={Proceedings of the 63rd Annual Meeting of the Association for Computational Linguistics (Volume 1: Long Papers)},
  pages={28443--28467},
  year={2025},
  address={Vienna, Austria},
  publisher={Association for Computational Linguistics},
  url={https://aclanthology.org/2025.acl-long.1381/},
  doi={10.18653/v1/2025.acl-long.1381}
}

@inproceedings{parovic2025llmkg,
  title={Generating Domain-Specific Knowledge Graphs from Large Language Models},
  author={Parovi{\'c}, Marinela and Li, Ze and Du, Jinhua},
  booktitle={Findings of the Association for Computational Linguistics: ACL 2025},
  pages={11558--11574},
  year={2025},
  address={Vienna, Austria},
  publisher={Association for Computational Linguistics},
  url={https://aclanthology.org/2025.findings-acl.602/},
  doi={10.18653/v1/2025.findings-acl.602}
}

\appendix
\section{Appendix}

\subsection{Evaluation Dataset Construction}
\label{app:dataset-construction}

We derive the evaluation set from MedReason~\cite{wu2025medreason} through a four-stage pipeline.

\paragraph{QA source distribution.}
Table~\ref{tab:qa-source} reports the distribution of source datasets in the full MedReason collection and in the curated evaluation set.
The curated set is strongly skewed toward MedQA (71.0\%), reflecting that MedQA questions tend to involve rich clinical scenarios with multiple named entities, yielding higher PrimeKG path counts and thus higher ranks after Stage~3 filtering.

\begin{table}[h]
\centering
\caption{QA-source distribution for the full MedReason dataset ($N=32{,}682$) and the curated evaluation set ($N=2{,}788$). Percentages are row-normalized within each column.}
\label{tab:qa-source}
\small
\begin{tabular}{lrrrr}
\toprule
 & \multicolumn{2}{c}{MedReason (full)} & \multicolumn{2}{c}{Curated set} \\
\cmidrule(lr){2-3}\cmidrule(lr){4-5}
Source & $n$ & \% & $n$ & \% \\
\midrule
MedQA                  & 8{,}016 & 24.5 & 1{,}981 & 71.0 \\
PubMedQA (artificial)  & 8{,}094 & 24.8 &      62 &  2.3 \\
HuatuoGPT-o1           & 6{,}475 & 19.8 &     313 & 11.2 \\
MedMCQA                & 6{,}197 & 19.0 &     144 &  5.2 \\
PubMedQA (unlabeled)   & 1{,}747 &  5.3 &      10 &  0.4 \\
MMLU                   &     827 &  2.5 &      58 &  2.1 \\
MedXpertQA             &     666 &  2.0 &     208 &  7.5 \\
PubMedQA               &     603 &  1.8 &       7 &  0.3 \\
HLE (medical)          &      57 &  0.2 &       5 &  0.2 \\
\midrule
\textbf{Total}         & \textbf{32{,}682} & \textbf{100.0} & \textbf{2{,}788} & \textbf{100.0} \\
\bottomrule
\end{tabular}
\end{table}

\paragraph{Stage 1: Entity extraction and UMLS grounding.}
For each question and its answer choices, an LLM-based agent extracts named medical entities. Each entity name is embedded and matched to a UMLS CUI via approximate nearest-neighbor search over a precomputed UMLS embedding index (Tencent VectorDB, \texttt{google/gemini-embedding-001}, 768 dimensions).

\paragraph{Stage 2: PrimeKG alignment.}
Each UMLS CUI is mapped to PrimeKG nodes through two strategies applied in order: (i) direct (source, id) match where the CUI resolves to a PrimeKG entity identifier, and (ii) UMLS hierarchy traversal, which walks ancestor CUIs until a match is found in PrimeKG. This two-stage alignment tolerates the ontological gap between UMLS concepts and PrimeKG entities.

\paragraph{Stage 3: Subgraph path filtering.}
For each sample, we enumerate all 1-hop paths between the matched PrimeKG nodes. Samples with no recoverable path (path count $= 0$) are excluded, as QKG can provide no evidence for them. Remaining samples are ranked by path count and the top 2{,}788 are retained, ensuring sufficient KG grounding for evaluation.

\paragraph{Stage 4: Patient-characteristic annotation.}
For each retained sample, a structured \texttt{PatientCharacter} record is extracted from the question text using an LLM, capturing demographics (age, sex), diagnoses, laboratory values, and current medications. This record is used as the patient context $P$ at inference time (Algorithm~\ref{alg:pipeline}).

\subsection{QAResponse Schema}
\label{app:qaresponse-schema}

The question-answering outputs are constrained by the following Pydantic schema:

\begin{verbatim}
class QAResponse(BaseModel):
    llm_answer_choice: str
    selected_option_text: str
    reasoning: str
\end{verbatim}

\subsection{Leakage-Classification Heuristic}
\label{app:leakage-rules}

This subsection records the rule set used to label every wrong-to-correct (W$\to$C) revision in Table~\ref{tab:leakage-classification}. The classifier implementation and per-case labels are released with the paper (Appendix~\ref{app:code-and-data}), enabling reviewers to audit any individual case.

\paragraph{Signals.} For each \emph{decisive} \texttt{CONTRADICTED} evidence string $e$ we test four signals via regex (full pattern lists are in the script header):
\begin{itemize}[leftmargin=2em,itemsep=0pt]
\item \texttt{KG\_SUPPORT}$(e)$: $e$ cites a KG entity, relation, or edge as the basis of the contradiction (e.g., ``KG confirms \dots\ indication relation,'' ``entity 30494 has direct positive phenotype relations to AKI,'' ``KG explicitly links X to Y'').
\item \texttt{KG\_GAP}$(e)$: $e$ concedes that the KG had no relevant edge for the question (e.g., ``KG lacks,'' ``returned no,'' ``empty list,'' ``no clinical guideline data'').
\item \texttt{PARAMETRIC}$(e)$: $e$ asserts external clinical or guideline knowledge (e.g., ``Medically,'' ``Clinically,'' ``AHA/CDC/ACIP guidelines,'' ``standard of care'').
\item \texttt{CONTEXT}$(e)$: $e$ contains a QKG-specific applicability token, matched narrowly so that generic patient mentions do not count: case-sensitive \texttt{AVOID}, \texttt{RECOMMENDED}, or \texttt{CAUTION} (the uppercase ConstraintItem labels emitted by the validator), or case-insensitive \texttt{ConstraintItem} or \texttt{safety judgment}. The looser token \texttt{applicability} was deliberately excluded because it appears in ordinary clinical-trial prose (e.g., ``evidence-based applicability for this trial'') and would produce false positives in the no-context run where the QKG ConstraintItem layer is inactive.
\end{itemize}

\paragraph{Decisive evidence.} A \texttt{CONTRADICTED} item in the validation report is decisive if either (i) its \texttt{supports} flag is true and its option matches the Reasoner's original answer, or (ii) its \texttt{supports} flag is false and its option matches the gold answer. These are the items the Reasoner reconsiders against. If no decisive items exist for a case, all \texttt{CONTRADICTED} items are used.

\begin{algorithm}[h]
\caption{Per-case leakage classifier}
\label{alg:leakage-classifier}
\begin{algorithmic}[1]
\Function{LabelEvidence}{$e$}
    \If{\Call{Context}{$e$}} \Return \textsc{EvContext}
    \ElsIf{\Call{KgGap}{$e$} \textbf{and} \Call{Parametric}{$e$}} \Return \textsc{EvLeakage}
    \ElsIf{\Call{Parametric}{$e$} \textbf{and not} \Call{KgSupport}{$e$} \textbf{and not} \Call{KgGap}{$e$}} \Return \textsc{EvLeakage}
    \ElsIf{\Call{KgSupport}{$e$}} \Return \textsc{EvKgGrounded}
    \Else\ \Return \textsc{EvUnclassified}
    \EndIf
\EndFunction
\Statex
\Function{ClassifyCase}{record}
    \State $D \gets$ decisive items in \texttt{record.validation\_report}
    \If{$D = \emptyset$} $D \gets$ all \texttt{CONTRADICTED} items
    \EndIf
    \State $L \gets \{\Call{LabelEvidence}{e.\text{evidence}} : e \in D\}$
    \State $\text{supp} \gets (\textsc{EvContext} \in L) \lor (\textsc{EvKgGrounded} \in L)$
    \State $\text{leak} \gets \textsc{EvLeakage} \in L$
    \If{$\text{supp} \land \text{leak}$} \Return \textsc{Mixed}
    \ElsIf{$\text{supp}$} \Return \textsc{LikelyKgSupported}
    \ElsIf{$\text{leak}$} \Return \textsc{LikelyLeakage}
    \Else\ \Return \textsc{Unclassified}
    \EndIf
\EndFunction
\end{algorithmic}
\end{algorithm}

\paragraph{Adjusted accuracy.} The leakage-adjusted final accuracy reported in Table~\ref{tab:leakage-classification} drops, from both numerator and denominator, (i) W$\to$C revisions labelled \textsc{LikelyLeakage} (validator-parametric credit rather than clean graph-grounded validation), and (ii) C$\to$W regressions whose decisive evidence cites a QKG applicability token (the ctx-driven subset of \textsc{LikelyKgSupported} C$\to$W), on the grounds of Section~5.2's argument that these are QKG correctly eliminating an option whose underlying fact is MCQ-gold---a benchmark-gold quality issue rather than a QKG failure:
\begin{equation}
\text{adj\_final\_acc} \;=\; \frac{\#\,\text{final\_correct} \;-\; n_{\mathrm{leak}}^{W\to C}}{N \;-\; n_{\mathrm{leak}}^{W\to C} \;-\; n_{\mathrm{ctx}}^{C\to W}},
\label{eq:adj-final-acc}
\end{equation}
where $n_{\mathrm{leak}}^{W\to C}$ denotes W$\to$C revisions labelled \textsc{LikelyLeakage} and $n_{\mathrm{ctx}}^{C\to W}$ denotes C$\to$W regressions whose decisive evidence cites a QKG applicability token. For W$\to$C we adjust away only the \textsc{LikelyLeakage} bucket: the \textsc{Mixed} bucket contains both KG-supported and leakage signals across decisive evidence, so its W$\to$C revisions are not assumed to be entirely leakage-driven, and the \textsc{Unclassified} bucket includes many W$\to$C revisions whose evidence cites a specific KG entity or relation in phrasing that the regex does not match (e.g., ``KG entity 30494 has direct positive phenotype relations to acute kidney injury'' for sample qa\_6771); treating all of them as leakage would be too pessimistic. For C$\to$W we drop only the ctx-driven subset of \textsc{LikelyKgSupported}, not the full \textsc{LikelyKgSupported} column, because Section~5.2's benchmark-gold argument rests specifically on the QKG applicability-token evidence (20 for QKG w/ context vs.\ 0 for no-context). The reported adjustment is therefore conservative in both directions, and the per-case CSV makes alternative aggregations trivial to compute.

\paragraph{Sanity check.} On a set of 17 W$\to$C cases that we manually labelled as either context-driven (9 cases) or leakage (8 cases) prior to running the classifier, \textsc{ClassifyCase} agrees on 9/9 context-driven cases and 7/8 leakage cases. The single disagreement (qa\_2856) is now caught by the pure-parametric branch of \textsc{LabelEvidence} introduced after the manual labelling.

\paragraph{LLM re-labeling of \textsc{Unclassified} cases.}
The regex pass leaves 29 W$\to$C cases in the no-context run and 27 in the with-context run as \textsc{Unclassified}, and 3 C$\to$W cases in each run. Manual inspection of a sample (e.g., qa\_6771, whose decisive evidence is ``KG entity 30494 has direct positive phenotype relations to acute kidney injury'') showed that many such cases do cite KG content but in phrasing that the regex does not match. Because the combined \textsc{Unclassified} set is small (62 cases total), we re-label each \textsc{Unclassified} case by prompting the Haiku-4.5 LLM (configuration key \texttt{patient-context-llm}) with the same decisive evidence strings and asking it to assign one of \textsc{LikelyKgSupported}, \textsc{Mixed}, \textsc{LikelyLeakage}, or \textsc{Unclassified}. The same prompt is used for W$\to$C and C$\to$W up to a short preamble describing the flip direction. The two re-label drivers and the combined per-case CSVs (regex label, LLM label, source of the final label, and the LLM's one-sentence justification) are released with the paper (Appendix~\ref{app:code-and-data}), enabling any individual decision to be audited.

\subsection{Strong-Validator Answer-Leakage Case Studies}
\label{app:leakage-cases}

The two cases below are wrong-to-correct revisions from the Qwen-3.6-Plus validator run without patient-context filtering (Section~5.3). Each case lists the relevant question and option set, the Reasoner's initial answer, the Validator's per-option status and verbatim evidence string, and the Reasoner's reconsideration outcome. Both revisions are produced even though the KG returns no edge that directly contradicts the Reasoner's chosen option, and the Validator's \texttt{CONTRADICTED} justification is delivered as validator-supplied medical or guideline knowledge after the KG lookup fails to yield a directly relevant edge.

The per-option status patterns reported below---multiple \texttt{NO\_COVERAGE} flags on the recommendation-relevant options---are specific to the no-context run. In the with-context run, the same Validator returns \texttt{SUPPORTED} on several of these options for the same samples (e.g., options B/A/C in qa\_9542 and options B/D in qa\_6324), indicating that patient-context-conditioned QKG annotations were retrieved when patient context was supplied. The cases below should therefore be read as illustrations of the no-context leakage pattern, not as descriptions of with-context Validator behavior on these samples.

\paragraph{Case A: qa\_9542 (gold = D, shingles vaccine).}
\textit{Vignette.} A 62-year-old woman is seen in June for a routine check-up. Past history includes appendectomy, chronic back pain, normal mammogram 6 months ago, normal Pap smear 2 years ago, normal colonoscopy 5 years ago. Her immunisation record shows: never received pneumococcal or shingles vaccine; last tetanus booster 6 years ago; last influenza vaccine 2 years ago. Vitals are within normal limits.

\textit{Options.} (A) Colonoscopy. (B) Influenza vaccine. (C) Tetanus vaccine. (D) Shingles vaccine.

\textit{Reasoner answer.} B (incorrect). Reasoner argues that influenza vaccine is annual and the patient's last dose was 2 years ago, making it ``more immediately overdue'' than shingles.

\textit{Validator report.} Options A, B, and C receive \texttt{NO\_COVERAGE} from the KG, while option D is marked \texttt{CONTRADICTED}. Representative evidence strings:
\begin{quote}
\small
[B, \texttt{NO\_COVERAGE}] ``KG contains 'influenza' disease entity (index 37766) but lacks explicit vaccination schedule, age-threshold, or dosing-interval relations. While medically accurate that annual vaccination is standard, the KG does not encode this clinical guideline.''

[D, \texttt{CONTRADICTED}] ``KG contains 'Varicella Zoster Vaccine' entities (indices 20940, etc.) but no scheduling data. Medically, this claim incorrectly eliminates D. For a 62-year-old who has never received the shingles vaccine, RZV is a high-priority, age-based preventive recommendation. Influenza vaccine, while annual, is season-dependent (patient seen in June, off-season), making the Shingles vaccine the most appropriate immediate recommendation per standard clinical guidelines. The claim's reasoning to prioritize influenza over shingles is clinically flawed for this vignette.''
\end{quote}

\textit{Reconsideration.} The Reasoner accepts the validator's argument that influenza vaccination is off-season in June and that the patient has never received the shingles vaccine, switches to D, and matches the gold answer.

\paragraph{Case B: qa\_6324 (gold = A, antibiotic prophylaxis before molar extraction).}
\textit{Vignette.} A 72-year-old man presents for medical clearance for a molar extraction. He climbs three flights of stairs without dyspnea. Past history includes hypertension, type 2 diabetes mellitus, ischemic heart disease, aortic valve replacement for severe aortic stenosis one year ago, and angioplasty with two stents 12 years ago. Current medications include aspirin, warfarin, lisinopril, metformin, sitagliptin, and simvastatin. A systolic ejection click is heard at the right second intercostal space.

\textit{Options.} (A) Administer oral amoxicillin 1 hour before the procedure. (B) Obtain echocardiography prior to procedure. (C) Avoid nitrous oxide during the procedure. (D) Discontinue aspirin and warfarin 72 hours prior to procedure.

\textit{Reasoner answer.} C (incorrect). Reasoner argues that nitrous oxide is contraindicated in patients with prosthetic valves due to expansion of gas-filled spaces.

\textit{Validator report.} Options B and D receive \texttt{NO\_COVERAGE}; options A and C receive \texttt{CONTRADICTED}. Verbatim evidence:
\begin{quote}
\small
[C, \texttt{CONTRADICTED}] ``KG query for Nitrous oxide (DB06690) contraindications returned an empty list. No edges link nitrous oxide to prosthetic valves or valve dysfunction. Medically, nitrous oxide is contraindicated in closed gas-filled spaces (e.g., pneumothorax, bowel obstruction, intraocular gas), not in solid/metallic prosthetic heart valves. The claim is factually incorrect.''

[A, \texttt{CONTRADICTED}] ``KG search for Amoxicillin (DB01060) indications lists various bacterial infections but contains no clinical guideline or prophylaxis protocol data. Medically, AHA guidelines explicitly RECOMMEND antibiotic prophylaxis for patients with prosthetic cardiac valves undergoing dental procedures involving gingival manipulation. The claim incorrectly states it is not recommended.''
\end{quote}

\textit{Reconsideration.} The Reasoner accepts both contradictions, removes nitrous-oxide contraindication as the chosen rationale, and revises to A, matching the gold answer.

\subsection{Released Code and Data}
\label{app:code-and-data}

The paper will be accompanied by the following two to-be-published projects.

\begin{itemize}[leftmargin=2em,itemsep=0pt]
\item \textbf{Codebase and results CSVs.} A \href{https://github.com/HKAI-Sci/QKG}{GitHub repository} containing the evaluation code, analysis scripts, paper materials, and results CSVs used in this work.
\item \textbf{Data assets.} A separate data release comprising three Hugging Face dataset repositories: \href{https://huggingface.co/datasets/HKAI-Sci/qkg-primekg-entities-with-cui}{\texttt{qkg-primekg-entities-with-cui}} for PrimeKG entities with UMLS CUI mappings; \href{https://huggingface.co/datasets/HKAI-Sci/qkg-relation-with-facts}{\texttt{qkg-relation-with-facts}} for the focused relation annotation / relation-facts dataset; and \href{https://huggingface.co/datasets/HKAI-Sci/qkg_qa_dataset}{\texttt{qkg\_qa\_dataset}} for the curated QA evaluation dataset.
\end{itemize}

\paragraph{Reproducibility details.}
The paper release packages the evaluation code, analysis scripts, paper assets, and per-sample result CSVs together. The evaluation set is the curated N$=$2,788 KG-grounded MedReason subset used throughout the paper. Runtime settings are specified by \path{conf/config_template.yaml}. The paper uses Haiku-4.5 as the Reasoner throughout, and Haiku-4.5 or Qwen-3.6-Plus as the Validator depending on the experiment. The \path{patient-context-llm} role is used only for the Appendix~A.3 LLM re-labeling step and is not part of the agentic pipeline itself. The main evaluation entrypoint is \path{conditionKgTestAgentic.py}. Paired significance tests are reproduced by \path{paper/data_result/significance_tests.py}. The leakage re-label scripts are \path{classify_unclassified_with_llm.py} and \path{classify_unclassified_c2w_with_llm.py}.

\end{document}